\documentclass[11pt,a4paper]{article}
\usepackage{authblk}

\usepackage{coling2020}

\usepackage[pdftex,bookmarks=true]{hyperref}
\hypersetup{colorlinks = false,  linkcolor = red,   anchorcolor = red,    citecolor = blue,  filecolor = red,   urlcolor = red}
\usepackage{times}
\usepackage{latexsym}
\usepackage{dirtytalk}
\usepackage{booktabs}

\usepackage{pgfplots}
\usepackage{standalone}
\usepackage{subcaption}

\usepackage[utf8]{inputenc}
\usepackage{url}
\usepackage{todonotes}
\usepackage{graphicx}
\usepackage{caption}
\usepackage{multirow}

\usepackage{bm}
\usepackage{amsmath}

\colingfinalcopy 

\newcommand{\emtext}[1]{{\em #1}}

\renewcommand{\vec}[1]{\mathbf{#1}}

\title{UWB at SemEval-2020 Task 1: Lexical Semantic Change Detection}

\DeclareMathOperator*{\argmin}{argmin~}

	\author[* 1,2]{\bf Ond\v{r}ej Pra\v{z}\'{a}k}
	\author[* 1,2]{\bf Pavel P\v{r}ib\'{a}\v{n}}
	\author[2]{\bf  Stephen Taylor}
	\author[1,2]{\bf Jakub Sido}
	\affil[1]{NTIS -- New Technologies for the Information Society,}
	\affil[2]{Department of Computer Science and Engineering,}
	\affil[ ]{Faculty of Applied Sciences, University of West Bohemia, Czech Republic}
	\affil[  ]{\tt	\{ondfa, pribanp, taylor, sidoj\}@kiv.zcu.cz}
	\affil[  ]{\tt	http://nlp.kiv.zcu.cz}

\date{}

\begin{document}

\maketitle
\begin{abstract}
In this paper, we describe our method for detection of lexical semantic change, i.e., word sense changes over time. We examine semantic differences between specific words in two corpora, chosen from different time periods, for English, German, Latin, and Swedish. Our method was created for the SemEval 2020 Task 1:  \textit{Unsupervised Lexical Semantic Change Detection.}  We ranked $1^{st}$ in Sub-task 1: binary change detection, and $4^{th}$ in Sub-task 2: ranked change detection. 
Our method is fully unsupervised and language independent. It consists of preparing a semantic vector space for each corpus, earlier and later; computing a linear transformation between earlier and later spaces, using Canonical Correlation Analysis and Orthogonal Transformation; 
and measuring the cosines between the transformed vector for the target word from the earlier corpus and the vector for the target word in the later corpus.


\end{abstract}
\section{Introduction}

%
%
\blfootnote{
    %
    %
    %
    %
    \hspace{-0.65cm}  
    ${}^{\text{*}}$Equal contribution.
    %
    %
}
\blfootnote{
    \hspace{-0.65cm}  
    This work is licensed under a Creative Commons 
    Attribution 4.0 International Licence.
    Licence details:
    \url{http://creativecommons.org/licenses/by/4.0/}.
}



Language evolves with time. New words appear, old words fall out of use, the meanings of some words shift.
The culture changes as well as the expected audience of the printed word.  There are changes in topics, in syntax, in presentation structure. Reading the natural philosophy musings of aristocratic amateurs from the 
eighteenth century, and comparing with a monograph from the nineteenth century, or a medical study from the twentieth century, we can observe differences in many dimensions, some of which seem hard to study. Changes in word senses are both a visible and a tractable part of language evolution.
Computational methods for researching words' stories have the potential of helping us understand this small corner of linguistic evolution. The tools for measuring these diachronic semantic shifts might also be useful for measuring whether the same word is  used in different ways in synchronic documents. The task of finding word sense changes over time is called diachronic \textit{Lexical Semantic Change (LSC)} detection. The task
is getting more attention in recent years \cite{hamilton-etal-2016-diachronic,frermann-lapata-2016-bayesian,schlechtweg-etal-2017-german}. There is also the \textit{synchronic} \textit{LSC} task, which aims to identify domain-specific changes of word senses compared to general-language usage \cite{schlechtweg-etal-2019-wind}.


 \newcite{tahmasebi2018survey} provides a comprehensive survey of  techniques for the \textit{LSC} task, as does \newcite{kutuzov-etal-2018-diachronic}. \newcite{schlechtweg-etal-2019-wind} evaluated available approaches for \textit{LSC} detection  using the \textit{DURel} dataset \cite{schlectweg-etal-DURel}.
 Some of the methodologies for finding time-sensitive meanings were borrowed from information retrieval techniques in the first place. According to \newcite{schlechtweg-etal-2019-wind}, there are mainly three types of approaches. (1) Semantic vector spaces approaches \cite{gulordava-baroni-2011-distributional,kim-etal-2014-temporal,Xu2015ACE,eger-mehler-2016-linearity,hamilton-etal-2016-cultural,hamilton-etal-2016-diachronic,rosenfeld-erk-2018-deep} represent each word with two vectors for two different time periods. The change of meaning is then measured by the cosine distance between the two vectors. (2) Topic modeling approaches \cite{wang-topics,bamman-topics,Wijaya-topic,mihalcea-nastase-2012-word,cook-etal-2014-novel,frermann-lapata-2016-bayesian,Schlechtweg20} estimate a probability distribution of words over their different senses, i.e., topics. (3) Clustering models \cite{mitra2015automatic,tahmasebi-risse-2017-finding} are used to cluster words into clusters representing different senses.

\par 
We participated in the \textit{SemEval 2020 Task 1: Unsupervised Lexical Semantic Change Detection} \cite{sem20-task1-overview} competition. In this paper, we describe our solution and submitted systems for this competition. 
The task consists of two sub-tasks, a \textit{binary classification} task (Sub-task 1) and a \textit{ranking} task (Sub-task 2), which involve comparing usage of target words between two lemmatized corpora, each drawn from documents from a different time period for four languages: \textit{English, German, Latin,} and \textit{Swedish}. For both sub-tasks, only the target words and two corpora for each language were provided by organizers, no annotated data. 
The task is intended to be solved in a completely unsupervised way.

\par In the \textit{binary classification} task, the goal is for two given corpora $C_1$ and $C_2$ (for time $t_1$ and $t_2$) and for a set of target words, decide which of these words changed or did not change their sense (semantic) between $t_1$ and $t_2$. \textit{Change of sense} is whether the word lost or gained any sense between the two periods (corpora).
The objective of the Sub-task 2 is for two given corpora $C_1$ and $C_2$, rank a set of target words according to their degree of lexical semantic change between $t_1$ and $t_2$.
A higher rank means a stronger change. Target words are the same for both sub-tasks.

Because language is evolving, 
expressions, words, and sentence constructions in two corpora from different time periods about the same topic 
will be written in languages that are quite similar but slightly different. They will 
share the 
majority of their words, grammar, and syntax. 

\par The main idea behind our solution is that we treat each pair of corpora $C_1$ and $C_2$ as different \textit{languages} $L_1$ and $L_2$ even though that text from both corpora is written in the same language.
We believe that these two languages $L_1$ and $L_2$ will be extremely similar in all aspects, including semantic. We train separate semantic space for each corpus and subsequently, we map these two spaces into one common cross-lingual space. We use methods for cross-lingual mapping \cite{Brychcin2019,artetxe-labaka-agirre:2016:EMNLP2016,artetxe-etal-2017,artetxe-etal-2018b,artetxe-etal-2018-robust} and thanks to the large similarity between $L_1$ and $L_2$ the quality of transformation should be high. We compute cosine similarity to classify and rank the target words, see Section \ref{sec:system-desc} for details. 

\nocite{Hardoon:2004}
\nocite{artetxe-etal-2018-robust}
\nocite{Hardoon:2004,artetxe-labaka-agirre:2016:EMNLP2016,FaruquiDyer14,Ammar2016,Brychcin2019}
\par Our systems\footnote{Our code is available at \url{https://github.com/pauli31/SemEval2020-task1}} ranked $1^{st}$ out of $33$ teams in Sub-task 1 with an average accuracy of $0.687$, and $4^{th}$ out of $32$ teams in Sub-task 2 with an average Spearman's rank correlation of $0.481$. 




\section{Data}
\par The corpora are drawn from several sources, described in \cite{sem20-task1-overview}.  Table \ref{tab1} shows periods and sizes.  For each language, items in the
earlier corpus are separated from items in the later one by at least 46 years (German) and as much as 2200 years (Latin). All corpora are lemmatized, punctuation is removed, and sentences are randomly reordered.
For English, target words have been marked with their part-of-speech. Two example sentences:
(1) \say{\textit{there be no pleasing any of you do as we may}}, and (2) \say{\textit{rise upon the instant in his stirrup the bold cavalier hurl with a sure and steady hand the discharge weapon in the face\_nn of his next opponent}}.
(1) illustrates a failure of lemmatization -- the word 'pleasing', which is a form of the verb 'please'; (2) shows a target word 'face', marked as a noun.  Less than 10\% of the 
English sentences contain a target word.

\begin{table*}[ht!]
\centering
\scalebox{0.85}{
\begin{tabular}{lrrrrr}
\toprule
{\bf         } &    \multicolumn{2}{c}{\textbf{Corpus 1}}  &\multicolumn{2}{c}{\textbf{Corpus 2}}& \\ \cmidrule(lr){2-3} \cmidrule(lr){4-5}
{\bf Language} &    \multicolumn{1}{c}{\textbf{Period}}& \multicolumn{1}{c}{\textbf{\#  Tokens}} &\multicolumn{1}{c}{\textbf{Period}}&  \multicolumn{1}{c}{\textbf{\# Tokens}}  &\textbf{\# Targets}\\ \cmidrule(lr){1-1} \cmidrule(lr){2-5} \cmidrule(lr){6-6} 
English&              1810-1860&          6 559 657             &1960-2010 & 6 769 814     &37\\
German&  1800-1900  & 70 244 484   &1946-1990 &72 397 546  &48 \\
Latin&200BC-1BC  &1 751 405 & 100AD-present & 9 417 033  &40   \\
Swedish& 1790-1830 & 71 091 465 &1895-1903 & 110 792 658         &31 \\

\bottomrule
\end{tabular}
}
\caption{Corpus statistics. The last column \textit{\# Targets} denotes the number of target words.}
\label{tab1}
\end{table*}





Lemmatization  reduces the vocabulary so that there are more examples of each  word.  It also introduces ambiguity; the decisions to add a POS tag to English target words and retain German noun capitalization shows that
the organizers were aware of this problem.

\section{System Description}
\label{sec:system-desc}

\par First, we train two semantic spaces from corpus $C_1$ and $C_2$. We represent the semantic spaces by a matrix $\vec{X}^s$ (i.e., a source space $s$) and a matrix $\vec{X}^t$ (i.e, a target space $t$)\footnote{The source space $\vec{X}^s$ is created from the corpus $C_1$ and the target space $\vec{X}^t$ is created from the corpus $C_2$.} using word2vec Skip-gram with negative sampling \cite{Mikolov2013a}. We perform a cross-lingual mapping of the two vector spaces, 
getting two matrices $\vec{\hat{X}}^s$ and $\vec{\hat{X}}^t$ projected into a shared space. We select two methods for the cross-lingual mapping \textit{Canonical Correlation Analysis (CCA)} using the implementation from \cite{Brychcin2019} and a modification of the \textit{Orthogonal Transformation} from \textit{VecMap} \cite{artetxe-etal-2018-robust}. Both of these methods are linear transformations. In our case, the transformation can be written as follows: 

\begin{equation}
  \vec{\hat{X}}^s =   \vec{W}^{s \rightarrow t} \vec{X}^s
\end{equation}
where $\vec{W}^{s \rightarrow t}$ is a matrix that performs linear transformation from the source space $s$ (matrix $\vec{X}^s$) into a target space $t$ and $\vec{\hat{X}}^s$ is the source space transformed into the target space $t$ (the matrix $\vec{X}^t$ does not have to be transformed because $\vec{X}^t$ is already in the target space $t$ and $\vec{X}^t = \vec{\hat{X}}^t$).

\par Generally, the CCA transformation transforms both spaces $\vec{X}^s$ and $\vec{X}^t$ into a third shared space  $o$ (where $\vec{X}^s\neq \vec{\hat{X}}^s$ and $\vec{X}^t\neq \vec{\hat{X}}^t$). Thus, CCA computes two transformation matrices $\vec{W}^{s \rightarrow o}$ for the source space and $\vec{W}^{t \rightarrow o}$ for the target space. The transformation matrices are computed by minimizing the negative correlation between the vectors $\vec{x}_{i}^{s} \in \vec{X}^s$ and $\vec{x}_{i}^{t} \in \vec{X}^t$ that are projected into the shared space $o$. The negative correlation is defined as follows:

\begin{equation}
    \argmin_{\vec{W}^{s \rightarrow o}, \vec{W}^{t \rightarrow o}} -\sum_{i=1}^{n} \rho (\vec{W}^{s \rightarrow o}\vec{x}_{i}^{s}, \vec{W}^{t \rightarrow o} \vec{x}_{i}^{t} ) = -\sum_{i=1}^{n}  \frac{cov(\vec{W}^{s \rightarrow o}\vec{x}_{i}^{s}, \vec{W}^{t \rightarrow o} \vec{x}_{i}^{t})}{\sqrt{var(\vec{W}^{s \rightarrow o}\vec{x}_{i}^{s}) \times var(\vec{W}^{t \rightarrow o} \vec{x}_{i}^{t})}}
\end{equation}
where $cov$ the covariance, $var$ is the variance and $n$ is a number of vectors. In our implementation of CCA, the matrix $\vec{\hat{X}}^t$ is equal to the matrix $\vec{X}^t$ because it transforms only the source space $s$ (matrix $\vec{X}^s$) into the target space $t$ from the common shared space with a pseudo-inversion, and the target space does not change. The matrix $\vec{W}^{s \rightarrow t}$ for this transformation is then given by:

\begin{equation}
     \vec{W}^{s \rightarrow t} = \vec{W}^{s \rightarrow o} (\vec{W}^{t \rightarrow o})^{-1}
\end{equation}

The submissions that use CCA are referred to as \textbf{cca-nn, cca-bin, cca-nn-r} and  \textbf{cca-bin-r} where the \textbf{-r} part means that the source and target spaces are reversed, see Section \ref{sec:exp-setup}. The \textbf{-nn} and \textbf{-bin} parts refer to a type of threshold used only in the Sub-task 1, see Section \ref{subsec:binary-system}. Thus, in Sub-task 2, there is no difference for the following pairs of submissions: \textbf{cca-nn} -- \textbf{cca-bin} and \textbf{cca-nn-r} -- \textbf{cca-bin-r}.

\par In the case of the \textit{Orthogonal Transformation}, the submissions are referred to as \textbf{ort} \& \textbf{uns}. We use Orthogonal Transformation with a supervised seed dictionary consisting of all words common to both semantic spaces. (\textbf{ort}). The transformation matrix $\vec{W}^{s \rightarrow t}$ is given by:

\begin{equation}
    \argmin_{\vec{W}^{s \rightarrow t}}{\sum_i^{|V|}{(\vec{W}^{s \rightarrow t} \vec{x}^s_i - \vec{x}^t_i)^2}}
\end{equation}
under the hard condition that $\vec{W}^{s \rightarrow t}$ needs to be orthogonal, where V is the vocabulary of correct word translations from source to target space. The reason for the orthogonality constraint is that linear transformation with the orthogonal matrix does not squeeze or re-scale the transformed space. It only rotates the space, thus it preserves most of the relationships of its elements (in our case it is important that orthogonal transformation preserves angles between the words, so it preserves the cosine similarity).

\newcite{artetxe-etal-2018-robust} also proposed a method for automatic dictionary induction. This is  a fully unsupervised method for finding orthogonal cross-lingual transformations. We used this approach for our \textbf{uns} submissions. 

Finally in all transformation methods, for each word $w_i$ from the set of target words $T$, we select its corresponding vectors $\vec{v}_{w_i}^{s}$ and $\vec{v}_{w_i}^{t}$ from matrices $\vec{\hat{X}}^s$ and $\vec{\hat{X}}^t$, respectively ($\vec{v}_{w_i}^{s} \in \vec{\hat{X}}^s$ and $\vec{v}_{w_i}^{t} \in \vec{\hat{X}}^t$), and we compute cosine similarity between these two vectors. The cosine similarity is then used to generate an output for each sub-task. For Sub-task 2, we compute the degree of change for word $w_i$ as $1 - cos (\vec{v}_{w_i}^{s}, \vec{v}_{w_i}^{t})$. The settings of hyper-parameters for both methods give us several combinations, see sections \ref{sec:exp-setup} and \ref{sec:results}.

\subsection{Binary System}
\label{subsec:binary-system}
The organizers provided a definition for binary change in terms of specific numbers of usages of senses of a target word.
We decided against attempting to model 
and group individual word usages.
%
Instead, we decided to use the continuous scores from Sub-task 2 
for the binary task.
We assume that there is a threshold $t$ for which the target words with a continuous score greater than $t$ changed meaning and words with the score lower than $t$ did not. We know that this assumption is generally wrong (because using the threshold we introduce some error into the classification), but we still believe it holds for most cases and it is the best choice. 
In order to examine
this assumption after the evaluation period, we computed the accuracy of the gold ranking scores with the optimal threshold (selected to maximize test accuracy). As you can see in Table \ref{tab:GoldAcc}, even if an optimal threshold is chosen,
the best accuracy that can be achieved is, on average, $87.6\%$.

\begin{table}[ht!]
    \centering
    \scalebox{0.78}{
    \begin{tabular}{cccccc}
    \toprule
         & \textbf{Avg} & \textbf{Eng} & \textbf{Ger} & \textbf{Lat} & \textbf{Swe}  \\ \midrule
         acc & .876 & .838 & .854 & .875 & .936 \\
         $t$ & - & .207 & .190 & .235 & .244 \\
         \bottomrule
    \end{tabular}
    }
    \caption{Sub-task 1 accuracy with gold ranking and optimal thresholds.}
    \label{tab:GoldAcc}
\end{table}

In order to find the threshold $t$, we tried several approaches. 
We call the first approach \textit{binary-threshold} (\textbf{-bin} in Table \ref{tab:rotatedResults}). For each target word $w_i$ we compute cosine similarity of its vectors $\vec{v}_{w_i}^{s}$ and $\vec{v}_{w_i}^{t}$, then we average these similarities for all words. The resulting averaged value is used as the threshold. Another approach called \textit{global-threshold} (\textbf{-gl}) is done similarly, but the average similarity is computed across all four languages. The last approach, called \textit{nearest-neighbors} (\textbf{-nn}), compares sets of nearest neighbours. For target word $w_i$ we find $100$ nearest (most similar) words from both transformed spaces (matrices $\vec{\hat{X}}^s$ and $\vec{\hat{X}}^t$), getting two sets of nearest neighbours $N^s$ and $N^t$ for each target word. Then, we compute the size of the intersection of these two sets for each target word. From the array of intersection sizes, we select the second highest value\footnote{We observed that the largest size of intersection is usually significantly greater than the other sizes so we select the second highest number. We are aware that this is a very simple and data-dependent approach and could be further improved.}, and we divide it by two. The resulting number is a threshold for all target words. If the size of a target word's intersection is greater or equal to the threshold, we classify the word \emtext{unchanged}; otherwise \emtext{changed}. This threshold is set for each language independently.



\section{Experimental Setup}
\label{sec:exp-setup}
To obtain the semantic spaces, we employ Skip-gram with negative sampling \cite{Mikolov2013a}. We use the Gensim framework \cite{rehurek_lrec} for training the semantic spaces. For the final submission, we trained the semantic spaces with $100$ dimensions for five 
iterations with five negative samples and window size set to five. Each word has to appear at least five times in the corpus to be used in the training. 
For all \textbf{cca-} submissions, we build the translation dictionary for the
cross-lingual transformation of the two spaces by removing the target words from the intersection of their vocabularies.

\par For \textbf{cca-nn-r} and \textbf{cca-bin-r} 
we change the direction of the cross-lingual transformation.
The initial setup for the transformation is that the source space is the space of the older corpus $C_1$ (represented by the matrix $\vec{X}^s$), and the target space is the semantic space of the later corpus $C_2$ (represented by the matrix $\vec{X}^t$). We reversed the order, and we use the matrix $\vec{X}^t$ as the source space, which is transformed into semantic space (matrix $\vec{X}^s$) of the older corpus, i.e. into the original source space. 

\par The two other methods are Orthogonal Transformations with identical words as the seed dictionary (\textbf{ort}) and unsupervised transformation (\textbf{uns}) from the \textit{VecMap} tool. In these methods, we use median similarity as the threshold for the binary task. We experimented with a separate threshold for each language (\textbf{bin} suffix) and with a single global threshold for all the languages (\textbf{gl} suffix).


\par The sample data provided with the task did not contain any labeled development target words, so we used the DURel \cite{schlectweg-etal-DURel} corpus and WOCC \cite{schlechtweg-etal-2019-wind} corpus\footnote{Both available at \url{www.ims.uni-stuttgart.de/en/research/resources/corpora/wocc}} to validate our solutions' performance. The WOCC corpus contains lemmatized German sentences from \textit{Deutsches Textarchiv} \cite{DTA}
for two time periods \textit{1750-1799} and \textit{1850-1899 }, which we used to train the semantic spaces. We evaluated our system on the DURel corpus, and we achieved $0.78$ of Spearman's correlation rank with settings that correspond to the column named \textbf{cca-nn-r} in Table \ref{tab:rotatedResults}.


\section{Results}
\label{sec:results}
\par We submitted eight different submissions for the Sub-task 1 and four for the Sub-task 2 obtained by CCA and the VecMap tool. We also submitted predictions based on Latent Dirichlet Allocation \cite{Blei03latentdirichlet} (LDA), but because of its poor results and limited paper size, we do not include the description here. The results, along with the ranking\footnote{The ranking in the table is the ranking among 186 submissions and does not  correspond to  ranking among 33 teams  because team ranks are based on their best submission.} are shown in Table \ref{tab:rotatedResults}. The bold results denote the best result we achieved for each language, and the underlined results were used in our team's final ranking. 


\begin{table}[h]

    \centering
    {
    \small
\begin{tabular}{lrcccccccccc}
\toprule
\bf Task& 
& \bf cca-nn & \bf cca-nn-r & \bf cca-bin & \bf cca-bin-r & \bf ort-bin & \bf ort-gl & \bf uns-bin & \bf uns-gl \\
\hline
\multirow{5}{*}{\bf 1 } &\bf  Avg & .572 (65) & .607 (28) & .663 (6) &\bf \textbf{\underline{0.687} (1)} & .634 (21) & .639 (16) & .659 (11) & .655 (13) \\
&\bf  Eng & .595 (6) &\bf .730 (1) & .595 (6) & .622 (5) & .568 (7) & .622 (5) & .568 (7) & .622 (5) \\
&\bf  Ger & .521 (14) & .542 (13) &\bf .812 (1) & .750 (3) & .750 (3) & .688 (6) & .750 (3) & .646 (8) \\
&\bf  Lat & .625 (5) & .575 (7) & .600 (6) & \textbf{.700 (2)} & .575 (7) & .600 (6) & .675 (3) & .675 (3) \\
&\bf  Swe & .548 (8) & .581 (7) & .645 (5) & \textbf{.677 (4)} & .645 (5) & .645 (5) & .645 (5) & .677 (4) \\
\hline
\multirow{5}{*}{\bf 2 } &\bf  Avg & .411 (14) & .469 (7) & - & - & \textbf{\underline{.481} (6)} & - & .455 (8) & - \\
&\bf  Eng & .279 (33) & \textbf{.421 (5)} & - & - & .367 (11) & - & .365 (12) & - \\
&\bf  Ger & \textbf{.707 (4)} & .696 (7) & - & - & .697 (6) & - & .687 (9) & - \\
&\bf  Lat & .181 (66) & .251 (58) & - & - & \textbf{.254 (55)} & - & .181 (65) & - \\
&\bf  Swe & .476 (10) & .509 (8) & -  & - &\bf .604 (1) &\bf - & .587 (3) & - \\
\hline
\end{tabular}}
    \caption{Results for our final submissions.}
    \label{tab:rotatedResults}
\end{table}

\par The \textit{cca-bin-r} system settings achieved the first-place rank out of 189 submissions by 33 teams on Sub-task 1, with an absolute accuracy of 0.687.  The majority class was \emtext{unchanged}, and always choosing it would have given a score of 0.571. 
The submissions with the top 15 best accuracies  (four teams) had scores ranging from 0.659 to 0.687 and assorted scores on the different languages.  
The four of our systems using  the \textbf{-bin} approach, which included the top-scoring system, have a mean percentile of 94.6; the two \textbf{-gl} strategy systems had a mean percentile of 92.7, and the two systems with the \textbf{-nn} strategy had a mean percentile of 75.5. On Sub-task 2, our best system achieved the fourth-place rank out of 33 teams.  It 
had an average correlation against the gold ranking for all four languages of 0.481.

Since the threshold strategy was used only for Sub-task 1, there is no difference in results for Sub-task 2  in Table \ref{tab:rotatedResults} in the following pairs of columns \textbf{cca-nn} -- \textbf{cca-bin}, \textbf{cca-nn-r} -- \textbf{cca-bin-r}, \textbf{ort-bin} -- \textbf{ort-gl} and \textbf{uns-bin} -- \textbf{uns-gl}. Thus, for Sub-task 2 we provide numbers only in the \textbf{cca-nn}, \textbf{cca-nn-r}, \textbf{ort-bin} and \textbf{uns-bin} columns.


Examining Table \ref{tab:rotatedResults}, the ranking scores for Latin are not only worse than the other
languages in absolute terms, but  their position relative to other submissions is also much worse.
The Latin corpora have several anomalies,
but only the small size of the earlier corpus (a third the size of the next smallest corpus) seems likely to be a problem.  For example, although both Latin corpora have a much larger proportion of 
short lines than the others, a measurement of the mean context size for target words shows that for all of the corpora, the mean context size for target words is between 8.38 (German 2) and 8.95 (German 1).

A different problem is syntax. The lemmatization of the corpora is no obstacle to the English reader, who can usually guess the correct form, because word-order in English (and also in Swedish and German) is somewhat rigid, and word inflexions are largely replaced by separate words, which have their own lemmas.  In Latin, word-order is quite flexible, and different authors have different preferred schemes for emphasis and euphony.  Two adjacent words are not required to be part of each other's context, and the most semantically related words in a sentence may be separated precisely to emphasize them.  

We performed post-evaluation experiments with word embedding vector size in order to discover its effect on system performance, see Figure \ref{fig:graphs} containing visualization of results for Sub-task 2. It shows that the optimal size for this task is in most cases between 100 and 175 for Sub-task 2, and between 75 and 125 for Sub-task 1 (not shown here). We also tried using fastText \cite{bojanowski2017enriching-fasttext} instead of word2vec Skip-gram to get the semantic space,  with settings that correspond to the ones we used for the final submissions. The performance using fastText was not improved in general, but for some  settings we obtained slightly better results. Other experiments suggest that Latin results can benefit when the context size is increased during the training of the semantic spaces. According to submitted results, it seems that CCA method with reversed order (columns \textbf{cca-nn-r},  \textbf{cca-bin-r} in Table \ref{tab:rotatedResults}) works better than without reversed but from the Figure \ref{fig:graphs} is evident that it is valid only for English and Latin.
\newcommand\xmin{25}
\newcommand\ymin{0.42}
\newcommand\ymax{0.83}

\newcommand\xminrank{25}
\newcommand\yminrank{-0.1}
\newcommand\ymaxrank{1.1}

\newcommand\heig{3.9cm}

\begin{figure*}[htb!]
	\begin{subfigure}{0.33\textwidth}
		\begin{center}
			\resizebox{\linewidth}{\heig}{%
				\pgfplotstableread[header=true]{data/SemEval-Method-a-rank-5-iter.tsv}\data

\begin{tikzpicture}
  \begin{axis}[
  title style={align=center,at={(0.5,-0.55)},anchor=north,yshift=-0.1},
  legend columns=4, 
   x label style={at={(axis description cs:0.5,-0.01)},anchor=north},
   xlabel={Embeddings dimension},
    legend pos = north east,
    xmin=\xminrank, xmax=300,
    ymin=\yminrank, ymax=\ymaxrank,
    xtick=data,
    ytick={0.0,0.1,0.2,0.3,0.4,0.5,0.6,0.7,0.8, 0.9, 1.0},
    x tick label style={rotate=55,anchor=east},
    ymajorgrids=true,
   grid style=dashed,
grid
    ]
    \addplot[blue, solid, every mark/.append style={solid},mark=triangle* ] table[x index = {0}, y index = {1}]{\data};
    \addplot[red, solid, every mark/.append style={solid}, mark=square*] table[x index = {0}, y index = {2}]{\data};
    \addplot[orange, solid, every mark/.append style={solid}, mark=pentagon*] table[x index = {0}, y index = {3}]{\data};
    \addplot[black, solid, every mark/.append style={solid}, mark=asterisk] table[x index = {0}, y index = {4}]{\data};
    \legend{Eng, Ger, Lat, Swe}
  \end{axis}
\end{tikzpicture}
			}
		\end{center}
		\caption{Nearest neighbors mapping -- \textit{cca-nn}}
		\label{fig:TODO4}
	\end{subfigure}\hspace*{\fill}
	\begin{subfigure}{0.33\textwidth}
		\begin{center}
			\resizebox{\linewidth}{\heig}{%
				\pgfplotstableread[header=true]{data/SemEval-Method-b-rank-5-iter.tsv}\data

\begin{tikzpicture}
  \begin{axis}[
  title style={align=center,at={(0.5,-0.55)},anchor=north,yshift=-0.1},
  legend columns=4, 
   x label style={at={(axis description cs:0.5,-0.01)},anchor=north},
   xlabel={Embeddings dimension},
    legend pos = north east,
    xmin=\xminrank, xmax=300,
    ymin=\yminrank, ymax=\ymaxrank,
    xtick=data,
    ytick={0.0,0.1,0.2,0.3,0.4,0.5,0.6,0.7,0.8, 0.9, 1.0},
    x tick label style={rotate=55,anchor=east},
    ymajorgrids=true,
   grid style=dashed,
grid
    ]
    \addplot[blue, solid, every mark/.append style={solid},mark=triangle* ] table[x index = {0}, y index = {1}]{\data};
    \addplot[red, solid, every mark/.append style={solid}, mark=square*] table[x index = {0}, y index = {2}]{\data};
    \addplot[orange, solid, every mark/.append style={solid}, mark=pentagon*] table[x index = {0}, y index = {3}]{\data};
    \addplot[black, solid, every mark/.append style={solid}, mark=asterisk] table[x index = {0}, y index = {4}]{\data};
    \legend{Eng, Ger, Lat, Swe}
  \end{axis}
\end{tikzpicture}
			}
		\end{center}
		\caption{Reversed nn. mapping -- \textit{cca-nn-r}}
		\label{fig:TODO5}
	\end{subfigure}\hspace*{\fill}
	\begin{subfigure}{0.33\textwidth}
		\begin{center}
			\resizebox{\linewidth}{\heig}{%
				\pgfplotstableread[header=true]{data/SemEval-Avg-rank-5-iter.tsv}\data

\begin{tikzpicture}
  \begin{axis}[
  title style={align=center,at={(0.5,-0.55)},anchor=north,yshift=-0.1},
  legend columns=2, 
   x label style={at={(axis description cs:0.5,-0.01)},anchor=north},
   xlabel={Embeddings dimension},
    legend pos = north east,
    xmin=\xminrank, xmax=300,
    ymin=\yminrank, ymax=\ymaxrank,
    xtick=data,
    ytick={0.0,0.1,0.2,0.3,0.4,0.5,0.6,0.7,0.8, 0.9, 1.0},
    x tick label style={rotate=55,anchor=east},
    ymajorgrids=true,
   grid style=dashed,
grid
    ]
    \addplot[blue, solid, every mark/.append style={solid},mark=triangle* ] table[x index = {0}, y index = {1}]{\data};
    \addplot[red, solid, every mark/.append style={solid}, mark=square*] table[x index = {0}, y index = {2}]{\data};
    \legend{\textit{cca--nn}, \textit{cca--nn--r}}
  \end{axis}
\end{tikzpicture}
			}
		\end{center}
		\caption{Average of all languages}
		\label{fig:TODO6}
	\end{subfigure}
	\caption{Performance using the CCA method with different word embeddings size for Sub-task 2.}
	\label{fig:graphs}
\end{figure*}
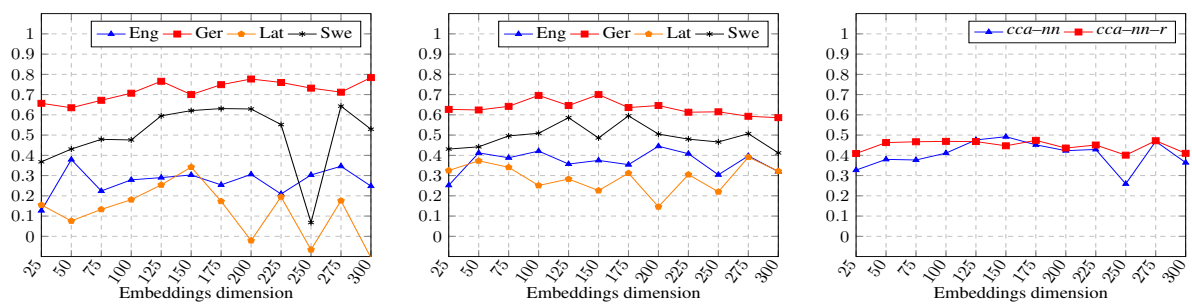

\vspace{-0.3cm}
\section{Conclusion}
Applying a threshold to semantic distance is a sensible architecture
for detecting the binary semantic change in target words between two corpora.  Our  \emtext{binary-threshold} strategy succeeded quite well.  We did have a small advantage in
that the list of target words turned out to be nearly equally divided between \emtext{changed} (0.57) and \emtext{unchanged} (0.43) words. Thus, choosing thresholds assuming that the division was 50/50 was not a severe problem.  Our experiments also reveal the limits of a threshold strategy, as shown in Table \ref{tab:GoldAcc}. Second, although our systems did not win Sub-task 2 change-ranking competition, they show that our architecture is a strong contender for this task when there is sufficient data to build good vector semantic spaces. The results for Latin illustrate that there is still room for other, less statistical approaches; one might have predicted, falsely, that 1.7M tokens was plenty of data.
The variety of experiments which were possible in the post-evaluation period suggest  that the corpora developed for the task will encourage work in the area.
In future work, we expect to focus on other cross-lingual techniques and other methods of measuring similarity between words besides cosine similarity.

 
\section*{Acknowledgments}
This work has been partly supported by ERDF ”Research and Development of Intelligent Components of Advanced Technologies for the Pilsen Metropolitan Area (InteCom)” (no.:   CZ.02.1.01/0.0/0.0/17 048/0007267), by the project LO1506 of the Czech Ministry of Education, Youth and Sports and by Grant No. SGS-2019-018 Processing of heterogeneous data and its specialized applications.  Access to computing and storage facilities owned by parties and projects contributing to the National Grid Infrastructure MetaCentrum provided under the programme "Projects of Large Research, Development, and Innovations Infrastructures" (CESNET LM2015042), is greatly appreciated.

\bibliography{sem20}

\begin{thebibliography}{}

\bibitem[\protect\citename{Ammar \bgroup et al.\egroup }2016]{Ammar2016}
Waleed Ammar, George Mulcaire, Yulia Tsvetkov, Guillaume Lample, Chris Dyer,
  and Noah~A. Smith.
\newblock 2016.
\newblock Massively multilingual word embeddings.
\newblock {\em Computing Research Repository}, arXiv:1602.01925.

\bibitem[\protect\citename{Artetxe \bgroup et al.\egroup
  }2016]{artetxe-labaka-agirre:2016:EMNLP2016}
Mikel Artetxe, Gorka Labaka, and Eneko Agirre.
\newblock 2016.
\newblock Learning principled bilingual mappings of word embeddings while
  preserving monolingual invariance.
\newblock In {\em Proceedings of the 2016 Conference on Empirical Methods in
  Natural Language Processing}, pages 2289--2294, Austin, Texas, November.
  Association for Computational Linguistics.

\bibitem[\protect\citename{Artetxe \bgroup et al.\egroup
  }2017]{artetxe-etal-2017}
Mikel Artetxe, Gorka Labaka, and Eneko Agirre.
\newblock 2017.
\newblock Learning bilingual word embeddings with (almost) no bilingual data.
\newblock In {\em Proceedings of the 55th Annual Meeting of the Association for
  Computational Linguistics (Volume 1: Long Papers)}, pages 451--462,
  Vancouver, Canada, July. Association for Computational Linguistics.

\bibitem[\protect\citename{Artetxe \bgroup et al.\egroup
  }2018a]{artetxe-etal-2018b}
Mikel Artetxe, Gorka Labaka, , and Eneko Agirre.
\newblock 2018a.
\newblock Generalizing and improving bilingual word embedding mappings with a
  multi-step framework of linear transformations.
\newblock In {\em Proceedings of the Thirty-Second AAAI Conference on
  Artificial Intelligence (AAAI-18)}, pages 5012--5019.

\bibitem[\protect\citename{Artetxe \bgroup et al.\egroup
  }2018b]{artetxe-etal-2018-robust}
Mikel Artetxe, Gorka Labaka, and Eneko Agirre.
\newblock 2018b.
\newblock A robust self-learning method for fully unsupervised cross-lingual
  mappings of word embeddings.
\newblock In {\em Proceedings of the 56th Annual Meeting of the Association for
  Computational Linguistics (Volume 1: Long Papers)}, pages 789--798,
  Melbourne, Australia, July. Association for Computational Linguistics.

\bibitem[\protect\citename{Bamman and Crane}2011]{bamman-topics}
David Bamman and Gregory Crane.
\newblock 2011.
\newblock Measuring historical word sense variation.
\newblock In {\em Proceedings of the 11th Annual International ACM/IEEE Joint
  Conference on Digital Libraries}, JCDL ’11, page 1–10, New York, NY, USA.
  Association for Computing Machinery.

\bibitem[\protect\citename{Blei \bgroup et al.\egroup
  }2003]{Blei03latentdirichlet}
David~M Blei, Andrew~Y Ng, and Michael~I Jordan.
\newblock 2003.
\newblock Latent dirichlet allocation.
\newblock {\em Journal of Machine Learning Research}, 3:2003.

\bibitem[\protect\citename{Bojanowski \bgroup et al.\egroup
  }2017]{bojanowski2017enriching-fasttext}
Piotr Bojanowski, Edouard Grave, Armand Joulin, and Tomas Mikolov.
\newblock 2017.
\newblock Enriching word vectors with subword information.
\newblock {\em Transactions of the Association for Computational Linguistics},
  5:135--146.

\bibitem[\protect\citename{Brychcín \bgroup et al.\egroup }2019]{Brychcin2019}
Tomáš Brychcín, Stephen Taylor, and Lukáš Svoboda.
\newblock 2019.
\newblock Cross-lingual word analogies using linear transformations between
  semantic spaces.
\newblock {\em Expert Systems with Applications}, 135:287--295.

\bibitem[\protect\citename{Cook \bgroup et al.\egroup
  }2014]{cook-etal-2014-novel}
Paul Cook, Jey~Han Lau, Diana McCarthy, and Timothy Baldwin.
\newblock 2014.
\newblock Novel word-sense identification.
\newblock In {\em Proceedings of {COLING} 2014, the 25th International
  Conference on Computational Linguistics: Technical Papers}, pages 1624--1635,
  Dublin, Ireland, August. Dublin City University and Association for
  Computational Linguistics.

\bibitem[\protect\citename{{Deutsches Textarchiv}}2017]{DTA}
{Deutsches Textarchiv}.
\newblock 2017.
\newblock Grundlage f\"ur ein referenzkorpus der neuhochdeutschen sprache.
\newblock Herausgegaben von der Berlin-Brandenburgischen Akademie der
  Wissenschaften. http://deutschestextarchiv.de.

\bibitem[\protect\citename{Eger and Mehler}2016]{eger-mehler-2016-linearity}
Steffen Eger and Alexander Mehler.
\newblock 2016.
\newblock On the linearity of semantic change: Investigating meaning variation
  via dynamic graph models.
\newblock In {\em Proceedings of the 54th Annual Meeting of the Association for
  Computational Linguistics (Volume 2: Short Papers)}, pages 52--58, Berlin,
  Germany, August. Association for Computational Linguistics.

\bibitem[\protect\citename{Faruqui and Dyer}2014]{FaruquiDyer14}
Manaal Faruqui and Chris Dyer.
\newblock 2014.
\newblock Improving vector space word representations using multilingual
  correlation.
\newblock In {\em Proceedings of the 14th Conference of the European Chapter of
  the Association for Computational Linguistics}, pages 462--471.

\bibitem[\protect\citename{Frermann and
  Lapata}2016]{frermann-lapata-2016-bayesian}
Lea Frermann and Mirella Lapata.
\newblock 2016.
\newblock A {B}ayesian model of diachronic meaning change.
\newblock {\em Transactions of the Association for Computational Linguistics},
  4:31--45.

\bibitem[\protect\citename{Gulordava and
  Baroni}2011]{gulordava-baroni-2011-distributional}
Kristina Gulordava and Marco Baroni.
\newblock 2011.
\newblock A distributional similarity approach to the detection of semantic
  change in the {G}oogle books ngram corpus.
\newblock In {\em Proceedings of the {GEMS} 2011 Workshop on {GE}ometrical
  Models of Natural Language Semantics}, pages 67--71, Edinburgh, UK, July.
  Association for Computational Linguistics.

\bibitem[\protect\citename{Hamilton \bgroup et al.\egroup
  }2016a]{hamilton-etal-2016-cultural}
William~L. Hamilton, Jure Leskovec, and Dan Jurafsky.
\newblock 2016a.
\newblock Cultural shift or linguistic drift? comparing two computational
  measures of semantic change.
\newblock In {\em Proceedings of the 2016 Conference on Empirical Methods in
  Natural Language Processing}, pages 2116--2121, Austin, Texas, November.
  Association for Computational Linguistics.

\bibitem[\protect\citename{Hamilton \bgroup et al.\egroup
  }2016b]{hamilton-etal-2016-diachronic}
William~L. Hamilton, Jure Leskovec, and Dan Jurafsky.
\newblock 2016b.
\newblock Diachronic word embeddings reveal statistical laws of semantic
  change.
\newblock In {\em Proceedings of the 54th Annual Meeting of the Association for
  Computational Linguistics (Volume 1: Long Papers)}, pages 1489--1501, Berlin,
  Germany, August. Association for Computational Linguistics.

\bibitem[\protect\citename{Hardoon \bgroup et al.\egroup }2004]{Hardoon:2004}
David~R. Hardoon, Sandor~R. Szedmak, and John~R. Shawe-Taylor.
\newblock 2004.
\newblock Canonical correlation analysis: An overview with application to
  learning methods.
\newblock {\em Neural Computation}, 16(12):2639--2664, December.

\bibitem[\protect\citename{Kim \bgroup et al.\egroup
  }2014]{kim-etal-2014-temporal}
Yoon Kim, Yi-I Chiu, Kentaro Hanaki, Darshan Hegde, and Slav Petrov.
\newblock 2014.
\newblock Temporal analysis of language through neural language models.
\newblock In {\em Proceedings of the {ACL} 2014 Workshop on Language
  Technologies and Computational Social Science}, pages 61--65, Baltimore, MD,
  USA, June. Association for Computational Linguistics.

\bibitem[\protect\citename{Kutuzov \bgroup et al.\egroup
  }2018]{kutuzov-etal-2018-diachronic}
Andrey Kutuzov, Lilja {\O}vrelid, Terrence Szymanski, and Erik Velldal.
\newblock 2018.
\newblock Diachronic word embeddings and semantic shifts: a survey.
\newblock In {\em Proceedings of the 27th International Conference on
  Computational Linguistics}, pages 1384--1397, Santa Fe, New Mexico, USA,
  August. Association for Computational Linguistics.

\bibitem[\protect\citename{Mihalcea and
  Nastase}2012]{mihalcea-nastase-2012-word}
Rada Mihalcea and Vivi Nastase.
\newblock 2012.
\newblock Word epoch disambiguation: Finding how words change over time.
\newblock In {\em Proceedings of the 50th Annual Meeting of the Association for
  Computational Linguistics (Volume 2: Short Papers)}, pages 259--263, Jeju
  Island, Korea, July. Association for Computational Linguistics.

\bibitem[\protect\citename{Mikolov \bgroup et al.\egroup }2013]{Mikolov2013a}
Tomas Mikolov, Kai Chen, Greg Corrado, and Jeffrey Dean.
\newblock 2013.
\newblock Efficient estimation of word representations in vector space.
\newblock In {\em Proceedings of workshop at ICLR}. arXiv.

\bibitem[\protect\citename{Mitra \bgroup et al.\egroup
  }2015]{mitra2015automatic}
Sunny Mitra, Ritwik Mitra, Suman~Kalyan Maity, Martin Riedl, Chris Biemann,
  Pawan Goyal, and Animesh Mukherjee.
\newblock 2015.
\newblock An automatic approach to identify word sense changes in text media
  across timescales.
\newblock {\em Natural Language Engineering}, 21(5):773--798.

\bibitem[\protect\citename{{\v R}eh{\r u}{\v r}ek and Sojka}2010]{rehurek_lrec}
Radim {\v R}eh{\r u}{\v r}ek and Petr Sojka.
\newblock 2010.
\newblock {Software Framework for Topic Modelling with Large Corpora}.
\newblock In {\em {Proceedings of the LREC 2010 Workshop on New Challenges for
  NLP Frameworks}}, pages 45--50, Valletta, Malta, May. ELRA.
\newblock \url{http://is.muni.cz/publication/884893/en}.

\bibitem[\protect\citename{Rosenfeld and Erk}2018]{rosenfeld-erk-2018-deep}
Alex Rosenfeld and Katrin Erk.
\newblock 2018.
\newblock Deep neural models of semantic shift.
\newblock In {\em Proceedings of the 2018 Conference of the North {A}merican
  Chapter of the Association for Computational Linguistics: Human Language
  Technologies, Volume 1 (Long Papers)}, pages 474--484, New Orleans,
  Louisiana, June. Association for Computational Linguistics.

\bibitem[\protect\citename{Schlechtweg and Walde}2020]{Schlechtweg20}
Dominik Schlechtweg and Sabine Schulte~im Walde.
\newblock 2020.
\newblock Simulating lexical semantic change from sense-annotated data.
\newblock In A.~Ravignani, C.~Barbieri, M.~Martins, M.~Flaherty, Y.~Jadoul,
  E.~Lattenkamp, H.~Little, K.~Mudd, and T.~Verhoef, editors, {\em The
  Evolution of Language: Proceedings of the 13th International Conference
  (EvoLang13)}.

\bibitem[\protect\citename{Schlechtweg \bgroup et al.\egroup
  }2017]{schlechtweg-etal-2017-german}
Dominik Schlechtweg, Stefanie Eckmann, Enrico Santus, Sabine Schulte~im Walde,
  and Daniel Hole.
\newblock 2017.
\newblock {G}erman in flux: Detecting metaphoric change via word entropy.
\newblock In {\em Proceedings of the 21st Conference on Computational Natural
  Language Learning ({C}o{NLL} 2017)}, pages 354--367, Vancouver, Canada,
  August. Association for Computational Linguistics.

\bibitem[\protect\citename{Schlechtweg \bgroup et al.\egroup
  }2018]{schlectweg-etal-DURel}
Dominink Schlechtweg, Sabine~Schulte im~Wlade, and Stefanie Eckmann.
\newblock 2018.
\newblock Diachronic usage relatedness (durel): A framework for the annotation
  of lexical semantic change.
\newblock In {\em Proceedings of NAACL-HLT 2018}, pages 169--174.

\bibitem[\protect\citename{Schlechtweg \bgroup et al.\egroup
  }2019]{schlechtweg-etal-2019-wind}
Dominik Schlechtweg, Anna H{\"a}tty, Marco Del~Tredici, and Sabine Schulte~im
  Walde.
\newblock 2019.
\newblock A wind of change: Detecting and evaluating lexical semantic change
  across times and domains.
\newblock In {\em Proceedings of the 57th Annual Meeting of the Association for
  Computational Linguistics}, pages 732--746, Florence, Italy, July.
  Association for Computational Linguistics.

\bibitem[\protect\citename{Schlechtweg \bgroup et al.\egroup
  }2020]{sem20-task1-overview}
Dominik Schlechtweg, Barbara McGillivray, Simon Hengchen, Haim Dubossarsky, and
  Nina Tahmasebi.
\newblock 2020.
\newblock {SemEval 2020 Task 1: Unsupervised Lexical Semantic Change
  Detection}.
\newblock In {\em Proceedings of the 14th International Workshop on Semantic
  Evaluation ({S}em{E}val-2020)}, Barcelona, Spain, Sep. Association for
  Computational Linguistics.

\bibitem[\protect\citename{Tahmasebi and
  Risse}2017]{tahmasebi-risse-2017-finding}
Nina Tahmasebi and Thomas Risse.
\newblock 2017.
\newblock Finding individual word sense changes and their delay in appearance.
\newblock In {\em Proceedings of the International Conference Recent Advances
  in Natural Language Processing, {RANLP} 2017}, pages 741--749, Varna,
  Bulgaria, September. INCOMA Ltd.

\bibitem[\protect\citename{Tahmasebi \bgroup et al.\egroup
  }2018]{tahmasebi2018survey}
Nina Tahmasebi, Lars Borin, and Adam Jatowt.
\newblock 2018.
\newblock Survey of computational approaches to lexical semantic change.
\newblock {\em arXiv preprint arXiv:1811.06278}.

\bibitem[\protect\citename{Wang and McCallum}2006]{wang-topics}
Xuerui Wang and Andrew McCallum.
\newblock 2006.
\newblock Topics over time: A non-markov continuous-time model of topical
  trends.
\newblock In {\em Proceedings of the 12th ACM SIGKDD International Conference
  on Knowledge Discovery and Data Mining}, KDD ’06, page 424–433, New York,
  NY, USA. Association for Computing Machinery.

\bibitem[\protect\citename{Wijaya and Yeniterzi}2011]{Wijaya-topic}
Derry~Tanti Wijaya and Reyyan Yeniterzi.
\newblock 2011.
\newblock Understanding semantic change of words over centuries.
\newblock In {\em Proceedings of the 2011 International Workshop on DETecting
  and Exploiting Cultural DiversiTy on the Social Web}, DETECT ’11, page
  35–40, New York, NY, USA. Association for Computing Machinery.

\bibitem[\protect\citename{Xu and Kemp}2015]{Xu2015ACE}
Yang Xu and Charles Kemp.
\newblock 2015.
\newblock A computational evaluation of two laws of semantic change.
\newblock In {\em CogSci}.

\end{thebibliography}
\bibliographystyle{coling.bst}

\end{document}